\documentclass{llncs}
\usepackage{amsmath,amssymb,amsbsy}
\usepackage{hyperref}       % hyperlinks

\let\epsilon\varepsilon
\let\phi\varphi

\def\N{\mathbb N}

\def\X{\mathcal X}
\def\x{{\mathbf x}}
\def\y{{\mathbf y}}

\def\E{E}

\begin{document}
%frontmatter 
%\pagestyle{headings}
 \title{Things Bayes can't do}
%\titlerunning{Things Bayes can't do}
 \author{Daniil Ryabko}
% \author{}
 %\author{This is an anonymized version of the paper [1] referenced in the submission. It is presented for the sake of completeness only; this is not submitted to NIPS}
% \date{}
\institute{INRIA \\ % 40 avenue de Halley, \\ 59650 Villeneuve d'Ascq, France\\ 
 \email{daniil@ryabko.net}}
\maketitle

 \begin{abstract}
  The problem of forecasting conditional probabilities of the next event given the past is considered
in a general probabilistic setting. Given an arbitrary  (large, uncountable) set $C$ of predictors, 
we would like to construct a single predictor that  performs asymptotically as well as  the 
best predictor in $C$, on any data.  Here we show that there are sets $C$ for which such predictors exist, 
but none of them is a Bayesian predictor with a prior  concentrated on $C$.
In other words, there is a predictor with sublinear regret, but every Bayesian predictor must have a linear regret. This negative finding is in sharp contrast with previous results
that establish the opposite  for the case when one of the predictors in $C$ achieves asymptotically vanishing error.
In such a case, if there is a predictor that achieves asymptotically vanishing error for any measure in $C$, then there is 
a Bayesian predictor that also has this property, and  whose prior is concentrated on (a countable subset of) $C$.
 \end{abstract}
%\mainmatter 
\section{Introduction}
The problem is probability forecasting in the most general  setting.
A sequence $x_1,\dots,x_t,\dots$ is generated by an unknown and arbitrary measure $\nu$ 
over the space of all infinite sequences.  Here for simplicity we consider $x_i$ coming 
from a finite set $\X$ (since we are after a negative result, this is not a limitation), but 
no other assumptions are made; in particular, $x_i$ may be dependent and the dependence may be arbitrary.
At each time step $t$ a predictor $\rho$ is required to give the conditional probabilities $\rho(x_{t+1}|x_1,\dots,x_t)$ of the next outcome $x_{t+1}$ given the observed past, before $x_{t+1}$ is revealed and the process continues. 
We would like  the predicted $\rho$ probabilities to be as close as possible to the unknown $\nu$ probabilities $\nu(x_{t+1}|x_1,\dots,x_t)$. The difference  is measured with respect to some loss function $L$, which in this work we take 
to be the $\nu$-expected average log loss (see the definitions below); however, it is clear that the main result applies more generally as well. 
Since $\rho$ is required to give conditional probabilities given every possible sequence of past outcomes, $\rho$ itself 
defines a probability measure over~$\X^\infty$, and thus predictors and environments (mechanisms generating the data) 
are objects of the same kind. 

To assist in the prediction task, we are given a set of predictors $C$. The performance of  our predictor $\rho$ is compared to
that of the predictors in $C$,  on sequences $x_1,\dots,x_t\dots$  generated by an arbitrary and unknown measure $\nu$. Thus, we are interested
in {\em  regret} of using  the predictor $\rho$ as opposed to using the best (for this~$\nu$) predictor from $C$.   The question 
we pose is whether this can be achieved by some kind of combination of predictors in $C$, or whether it may be necessary 
to look elsewhere~--- outside  of~$C$ (and  its convex hull).   More specifically, we are asking the question 
of {\em whether there exists a prior over  $C$, such that the Bayesian predictor with this prior has the smallest 
possible regret (at least,  asymptotically) with respect to the best measure in $C$}.  The answer we obtain is negative: there are some 
classes $C$ such that any Bayesian predictor has linear regret, while  the best possible predictor has sublinear regret.
Note that an example of such a set $C$ is {\em necessarily uncountable}, since for a countable set $C$ 
any prior with non-zero weights results in a Bayesian predictor with at most constant (in time) regret with respect to each 
predictor in $C$, and thus zero asymptotic average regret. 

It is worth noting that the result is not about Bayesian versus non-Bayesian inference; in fact, in the last section of the paper it is argued 
 that the negative finding applies  not only to  Bayesian predictors. Thus, 
the result means that in some cases, given a set of predictors, to construct a predictor that performs as close as possible
to the best  of them, one has to look elsewhere~--- somewhere completely outside of~$C$ (and its convex hull).

\noindent{\bf Prior work.} This is somewhat disturbing, since it contradicts both the intuition acquired 
from the literature on less general cases, and the positive results in related general settings. 
Specifically, the question has been studied extensively for specific families  $C$ of predictors, 
as well as in  the non-probabilistic setting of prediction with expert advice. For specific families, 
the question dates back to Laplace who considered it for the case when $C$ is the set of all Bernoulli i.i.d.\ 
measures, and, moreover, it is assumed that the measure $\nu$ to be predicted belongs to $C$. The latter assumption 
means that the problem is in the {\em realizable case}. The predictor suggested by Laplace is in fact a Bayesian predictor 
with the uniform prior over the parameter space. Moreover, a Bayesian predictor (with a different, Dirichlet, prior) 
is known to achieve optimal cumulative log loss in the realizable case of this problem, and, more generally for the case when 
$C$ is the set of Markov processes of order $k$ \cite{Krichevsky:93}.  
Bayesian predictors for a variety of other families are widely used, and their optimality can be often established 
even outside the Bayesian setting, including the settings where the measures to be predicted are outside the  predictor's prior. 
For example, a Bayes mixture over all finite-memory processes predicts also all stationary processes \cite{BRyabko:88}.

 In the setting  of prediction with expert advice, one is given  a finite 
set $C$ of experts, and the predictor that competes with them  is constructed that has a small regret (see, e.g., \cite{Cesa:06} for an extensive overview). 
%This regret  is constant in the log loss setting (just as is the case with the Bayes mixture over a finite set).
 A typical construction for such a 
 combination of experts is obtained by attaching a weight to each expert's prediction, where the weight decreases
exponentially with the  loss accumulated~--- a construction that is clearly reminiscent of Bayesian updating. 

 In either setting, one is typically concerned with finite or countable classes, 
or with some specific parametric families of experts.
The general case of the prediction problem has been formulated in \cite{Ryabko:10pq3+}, where it is shown that, if we are only interested 
in the realizable case, that is, the measure $\nu$ to be predicted belongs to $C$, then one can always do with a Bayes
predictor. More precisely, if there is a predictor $\rho$ whose error  asymptotically vanishes with $t$ on every
$\nu\in C$, then there is a Bayesian predictor (with a prior over a measurable subset of $C$) that also has this property.
Moreover, the prior can be always taken over a countable subset of $C$. This is shown without any assumptions on $C$ whatsoever;
in particular, $C$ is not required to be measurable. The work  \cite{Ryabko:11pq4+} unifies the formulations of the realizable 
and the non-realizable (expert advice) problems, and also formulates the following semi-realizable problem to which 
the result of \cite{Ryabko:10pq3+} is generalized: 
 now $\nu$ is allowed to be any measure such that there is a measure $\mu$ in $C$ 
whose error asymptotically vanishes on $\nu$. Here, again, if anything works then there is a prior such that a Bayesian
predictor with this prior works as well.  The present work completes the picture (and answers an open question from \cite{Ryabko:11pq4+}), showing that, unlike 
the realizable and semi-realizable case, the fully non-realizable case of the problem cannot always be solved 
by a Bayesian predictor. 

The result of this work   along with those cited above can be also put into the perspective of classical results 
on the consistency of Bayesian inference. Thus,  in  \cite{Diaconis:86} it is shown that, roughly speaking, 
  there may exist a prior with which a Bayesian predictor is inconsistent. In the context of the realizable case
of the prediction problem, that is, if there is a consistent predictor,  \cite{Ryabko:10pq3+} shows that there always exists a prior with which a Bayesian predictor is consistent.
Here we show that, in the nonrealizable case, there are cases where every Bayesian predictor with {\em every possible prior}
 is far from being as close as possible to being consistent.

\section{Preliminaries}
Let $\X$ be  a finite set (the alphabet). Denote $\X^*:=\cup_{k\in\N}\X^k$. The notation $x_{1..T}$ is used for $x_1,\dots,x_T$. 
 We consider  stochastic processes (probability measures) on $(\X^\infty,\mathcal B)$ where $\mathcal B$
is the usual Borel sigma-field. 
%  generated by the cylinder sets  $[x_{1..T}]$, $x_i\in\X, T\in\N$ 
% and $[x_{1..T}]$ is the set of all infinite sequences that start with $x_{1..T}$.
% For a  finite set $A$ denote $|A|$ its cardinality.
We use  $\E_\mu$ for
expectation with respect to a measure $\mu$.

The loss we use in this paper is the expected log loss, which can be defined as 
 the  expected cumulative Kullback-Leibler divergence (KL divergence): 
\begin{equation*} 
  L_T(\nu,\rho):=  \E_\nu
  \sum_{t=1}^T  \sum_{a\in\X} \nu(x_{t}=a|x_{1..t-1}) \log \frac{\nu(x_{t}=a|x_{1..t-1})}{\rho(x_{t}=a|x_{1..t-1})},
\end{equation*}
where $\nu,\rho$ are any measures over $\X^\infty$.
In words, we take the expected (over data) cumulative (over time) KL divergence between $\nu$- and $\rho$-conditional (on the past data) 
probability distributions of the next outcome. 
The expected  log loss is easy to study because of the following identity
\begin{equation}\label{eq:kl}
 L_T(\nu,\rho)=-E_\nu \log \frac{\rho(x_{1..T})}{\nu(x_{1..T})},
\end{equation}
where on the right-hand side we have simply the KL divergence between measures $\mu$ and $\rho$ restricted to the first $T$ observations.

If we have two predictors $\mu$ and $\rho$, we can  define the {\em regret} up to time $T$ of 
(using the predictor) $\rho$ as opposed to (using the predictor) $\mu$ on the measure $\nu$ (that is, $\nu$ generates the sequence to predict) as 
$$
R_T^\nu(\mu,\rho):=L_T(\nu,\rho) - L_T(\nu,\mu).
$$

For a set of measures $C$  one can also define the regret up to time $T$ of 
$\rho$ with respect to $C$ on $\nu$ as 
$
R_T^\nu(C,\rho):=\sup_{\mu\in C} R_T^\nu(\mu,\rho).
$
For the case of a finite or compact $C$ one often seeks to minimize $R_T^\nu(C,\rho)$. However, 
already for countably infinite sets $C$ it may not be possible to bound $R_T^\nu(\mu,\rho)$ uniformly over $C$.
 This is why we will not make much use of $R_T^\nu(C,\rho)$,  but rather work 
with its asymptotic version, defined as follows. 

Define the asymptotic average regret as 
$$
\bar R^\nu(\mu,\rho):=\limsup_{T\to\infty}{1\over T}R^\nu_T(\mu,\rho),
$$
and
$$
\bar R^\nu(C,\rho):= \sup_{\mu\in C}\bar R^\nu(\mu,\rho).
$$
Note that, since we are after a negative result, working with asymptotic quantities only is not a limitation.

\section{Main result}
\begin{theorem}
 There exist a set $C$ of measures and a predictor $\rho$ such that for every measure $\nu$ we have
%$R_T^\nu(C,\rho)\le cT +o(T)$, 
% $\bar R^\nu(C,\rho)\le c$ for every measure $\nu$, yet
 $\bar R^\nu(C,\rho) = 0 $, yet
for every Bayesian predictor $\phi$ with a prior concentrated on $C$ 
there exists a measure $\nu$ such that  % $R_T^\nu(C,\phi)\ge 2cT + o(T)$ and 
 $\bar R^\nu(C,\phi)\ge c>0$ 
where $c$ is a (possibly large) constant.
In other words, any  Bayesian predictor must have a linear regret, while there exists a predictor with a sublinear regret.
%That is, the loss of any Bayesian predictor  is about double the loss of~$\rho$.
\end{theorem}
\begin{remark}[Countable $C$]\label{r:count}
Note that any set $C$ satisfying the theorem must necessarily be uncountable.
Indeed, for any countable set $C=(\mu_k)_{k\in\N}$, take the Bayesian predictor $\phi:=\sum_{k\in\N}w_k\mu_k$,
where $w_k$ can be, for example, $\frac{1}{k(k+1)}$. Then, for any $\nu$ and any $T$, from~\eqref{eq:kl} we obtain
$$L_T(\nu,\phi)\le-\log w_k + L_T(\nu,\mu_k).$$
That is to say, the regret of $\phi$ with respect to any $\mu_k$ is a constant independent of $T$ (though it does depend on $k$), and thus for every $\nu$ we have
$\bar R^\nu(C,\phi)=0$.
It is worth noting that the origins of the use  of such countable mixtures for prediction trace back to \cite{Zvonkin:70,Solomonoff:78}.
\end{remark}

Before passing to the proof of the theorem, we present here an informal exposition of the 
counterexample used in the proof and the idea why it works.

The example of the proof starts with taking a Bernoulli i.i.d.\ biased coin-toss measure, say, the one  with the parameter $p=1/3,$
denoted $\beta_p$. 
Take then the set $S$ of sequences typical for this measure, that is, all sequences for which the frequency of 1
is asymptotically 1/3.
 We are interested in a predictor that  predicts all measures concentrated on a single sequence from $S$, and we will  ignore all other possible $\nu$.
 The set  of measures $C$ is constructed as follows. Take any sequence $\x$ in $S$ and 
define the measure $\mu_\x$ as the one that behaves exactly as Bernoulli 1/3 on this sequence $\x$, and on all 
other sequences it behaves as some fixed (deterministic) measure.
 In other words, we have taken a Bernoulli 1/3 measure and split it into all its typical sequences,
continuing it with a fixed arbitrary sequence everywhere else. Denote $C$ the resulting set of measures. Note that the original measure $\beta_p$
can be recovered with a Bayesian predictor from the set $S$. Indeed, it is enough to take $\beta_p$ itself as a prior 
over $S$. Such a Baysian predictor will then be as good as $\beta_p$ on any measure. Observe that for every $x_{1..T}$ it puts
the  weight of about $2^{-h_p T}$  on the set of sequences from $S$ that start with $x_{1..T}$  (where $h_p$ is the binary entropy for $p=1/3$ of the example). The loss it achieves on 
measures from $S$ is thus $h_p T$ and this is,  in fact, also  the minimax 
 loss one can achieve on   $S$.   However, it is not possible 
to achieve the same loss (and to recover $\beta_p$) with a Bayesian predictor whose prior is concentrated on the set~$C$. The trouble is that each measure $\mu$ in $C$
attaches already too little weight to the sequence from $S$ that  it is based on. To be precise, the weight it attaches
is the same $2^{-h_p T}$ that the Bayesian predictor gives to the corresponding deterministic sequence. Whatever extra prior weight a  Bayesian 
predictor gives will only go towards regret; it cannot give a constant weight to each measure because there are uncountably many 
of them. In fact, the best it can do is give another $2^{-h_p T}$,  which means that the resulting loss is going to be double  the best possible
one can obtain on measures from $S$ with the best possible predictor, and, again, double of what one can obtain taking 
for each $\nu\in S$ the best $\mu\in C$. This results in linear regret, which is, as we show, is at least $h_p$ in asymptotic average.

\begin{proof}
Let the alphabet $\X$  be  ternary $\X=\{0,1,2\}$. 
 For $\alpha\in(0,1)$ denote $h(\alpha)$ the binary entropy $h(\alpha):= -\alpha\log \alpha-(1-\alpha)\log(1-\alpha)$.
Fix an arbitrary  $p\in(0,1/2)$  %be defined as $h^{-1}(1/2)$ 
 and let $\beta_p$ be the Bernoulli i.i.d.\ measure (produces only 0s and 1s) with parameter $p$. 
Let $S$ be the set of sequences in $\X^\infty$ that have no $2$s and such that the frequency of $1$ is close to $p$:
\begin{multline*}
 S:=\{{\bf x}\in\X^\infty:  x_i\ne2 \forall i, \text { and} \\   \left|{1\over t}|\{i=1..t:x_i=1\}| - p\right|\le f(t) \text{ from some $t$ on}\},
\end{multline*}
where $f(t)=\log t/\sqrt{t}$. Clearly, $\beta_p(S)=1$. 

 Define the set $D_S$ as the set of all Dirac measures concentrated on a sequence
from $S$, that is $D_S:=\{\nu_\x: \nu_\x(\x)= 1,\ \x\in S\}$. Moreover, for each $\x\in S$ define the measure $\mu_\x$ as follows: 
$\mu_\x(X_{T+1}|X_{1..T})=p$ coincides with $\beta_p$ (that is, 1 w.p. $p$ and 0 w.p. $1-p$) if $X_{1..T}=x_{1..T}$, 
 and outputs 2 w.p.~1 otherwise: $\mu_\x(2|X_{1..T})=1$ if $X_{1..T}\ne x_{1..T}$.  That is, $\mu_\x$ behaves
as $\beta_p$ only on the sequence $\x$, and on all other sequences it just outputs 2 deterministically. 
This  means, in particular, that many sequences have probability 0, and some probabilities above are defined conditionally on 
zero-probability events,  but this is not a problem; see the remark in the end of the proof. 

Finally, let $C:=\{\mu_\x: \x\in S\}$. Next we will define the predictor $\rho$ that predicts 
well all measures in $C$. First, introduce the measure $\delta$ that is going to take care of 
all the measures that output 2 w.p.1 from some time on. For each $a\in\X^*$ let $\delta_a$ be the 
measure that is concentrated on the sequence that starts with $a$ and then consists of all 2s.
Define $\delta:=\sum_{a\in\X^*}w_a\delta_a$, where $w_a$ are arbitrary positive numbers that sum to 1. 
Let also the measure  $\beta'$ be  i.i.d.\ uniform over $\X$. 
 Finally, define    
\begin{equation}\label{eq:3}
\rho:=1/3(\beta_p+\beta'+\delta).
\end{equation}

Next, let us show that, for every $\nu$, the measure  $\rho$ predicts $\nu$ as well as any measure in $C$:
its loss is an additive constant factor. 
In fact, it is enough to see this for all $\nu\in D_S$, and for all measures that output all 2s w.p.1\ from some $n$ on.
For each $\nu$ in the latter set, from~\eqref{eq:3} the loss of $\rho$ is upper-bounded by $\log 3  -\log {w_a}$, where $w_a$ is 
the corresponding weight. This is a constant (does not depend on $T$). For the former set, again from the definition~\eqref{eq:3}
for every $\nu_\x\in D_S$  we have  (see also Remark~\ref{r:count})
$$
L_T(\nu_\x,\rho)\le \log 3 + L_T(\nu_\x,\beta_p)= Th_p+o(T),$$  while 
$$\inf_{\mu\in C} L_T(\nu_\x,\mu)= L_T(\nu_\x,\mu_\x) = Th_p+o(T).$$ 
Therefore, for all $\nu$ we have 
$$
 R^\nu_T(C,\rho)=o(T)\text{ and }\bar R^\nu(C,\rho)=0.
$$
Thus, we have shown that for 
every $\nu\in S$ there is a reasonably good predictor in $C$ (here ``reasonably good'' means
that its loss is linearly far from  that of random guessing), and, moreover, there is a predictor
$\rho$ whose asymptotic regret is zero with respect to $C$. 

Next we need to show that any Bayes predictor has $2Th_p + o(T)$ loss on at least some measure, which 
is double that of $\rho$, and which can be as bad as random guessing  (or worse; depending on $p$).
 We will 
show something stronger:  any Bayes predictor has asymptotic  average loss  of $2Th_p$ {\em on average} 
over all measures in $S$. So there will be many measures on which it is bad, not just one. 

  Let $\phi$ be any Bayesian predictor with its prior concentrated on  $C$. Since $C$ is parametrized by $S$,
for any $x_{1..T}\in\X^T, T\in\N$ we can write  $\phi(x_{1..T})= \int_S \mu_\y(x_{1..T})dW(\y)$
where $W$ is some measure  over $S$ (the prior).
Moreover, using the notation $W(x_{1..k})$ for the $W$-measure of all sequences in $S$ that 
start with $x_{1..k}$, from the definition of the measures $\mu_\x$, for every $\x\in S$  we have  
\begin{equation}\label{eq:unint}
 \int_S \mu_\y(x_{1..T})dW(\y)  = \int_{{\bf y}\in S: y_{1..T}=x_{1..T}} \beta_p(x_{1..T})dW(\y) =  \beta_p(x_{1..T}) W(x_{1..T}).
\end{equation}

 We will consider 
the average 
$$E_U \limsup {1\over T} L_T(\nu_x,\phi) dU(\x),$$
 where the expectation is taken with respect to 
the measure $U$ defined as the measure $\beta_p$ 
 restricted to $S$; in other words,  $U$ is approximately   uniform  over this set.
Fix any $\nu_\x\in S$. Observe that  $L_T(\nu_{\bf x},\phi)=-\log\phi(x_{1..T})$.
For the asymptotic regret, we can assume w.l.o.g.\  that the loss  $L_T(\nu_\x,\phi)$  is upper-bounded, say,
 by $T\log|\X|$ at least from some $T$ on
(for otherwise the statement already holds for $\phi$). This allows us to 
 use  Fatou's lemma to bound 
\begin{multline}\label{eq:f}
 E_U \limsup {1\over T} L_T(\nu_\x,\phi) \\ \ge  \limsup {1\over T} E_U  L_T(\nu_\x,\phi)  
=    \limsup - {1\over T}  E_U \log\phi(\x) \\ = \limsup - {1\over T}  E_U \log  \beta_p(x_{1..T}) W(x_{1..T}),
\end{multline} 
where in the last equality we used~\eqref{eq:unint}. 
%Thus, we can deal with time-$T$ loss (and regret) further on.
Moreover, 
\begin{multline}
- E_U \log \beta_p(x_{1..T}) W(x_{1..T})\\= -  E_U \log \beta_p(x_{1..T})  + E_U \log \frac{U(x_{1..T})}{ W(x_{1..T})} - E_U \log  U(x_{1..T}) \ge 2h_pT + o(T),
\end{multline}
where in the inequality we have used the fact that KL divergence is non-negative and the definition of $U$ (that is, that $U=\beta_p|_S$).
From this and~\eqref{eq:f} we obtain the statement of the theorem.

Finally, we remark that all the considered measures can be made non-zero everywhere by simply combining them with 
the uniform i.i.d.\  over  $\X$ measure $\beta'$, that is, taking for each measure $\nu$ the combination ${1\over 2}(\nu +\beta')$.
This way all losses up to time $T$ become bounded by $T\log|\X|+1$, but the result still holds with a different constant.
\qed\end{proof}

\section{Discussion}
We have shown that there are sets of predictors whose performance cannot be combined using any Bayesian 
predictor. While the result is stated for Bayesian predictors and for log loss, 
the  example used to  establish it seems to apply more generally. Indeed, it is clear that changing the loss
won't change the result, only making  the analysis slightly more cumbersome. More generally, the reason why any Bayesian 
predictor does not work in this example is that, since the set $C$ considered is large, the predictor has to attach a quickly decreasing
weight to each element in $C$, whereas each measure in $C$ already attaches too little  weight to the part 
of the event space of interest. In other words, the likelihood of the observations w.r.t.\ each predictor in $C$ is too small
to allow for any penalty. To combine predictors in $C$ one has to {\em boost} the likelihood, rather than attach a penalty.
Doing something like this would of course break a predictor on  other sets~$C$. This applies not only to Bayesian prediction.
In fact, whatever general prediction principle one could consider, for example, the MDL principle (see, e.g., \cite{Grunwald:07}),
it appears to fail  on the example presented. The same concerns expert-advice-style predictors. 
  The problem, therefore, seems to be generic: to combine the predictive power of the 
predictors in the set, it is not enough to consider combinations of these predictors; rather, one
has to look somewhere completely outside of $C$. 

 This suggests a more general question of how  one can characterize those sets $C$  of predictors 
for which it is enough to consider only the predictors inside $C$ in order to effectively compete with them, 
as well as what can one do when this is not the case.

As far as prediction in the realizable case is concerned (i.e., $\nu\in C$, or, more generally, zero regret is possible), 
the following question remains open. It is shown in \cite{Ryabko:10pq3+} that in this case, if any predictor 
works then there is a prior such that a Bayesian predictor with this prior works as well. However, this result is asymptotic.
One can ask the question of whether the speed of convergence of the loss (to 0 in this case) can be matched by some 
Baysian predictor, if any of the measures in $C$ is chosen to generate the data. 

Another generalization is to the case when the best achievable regret is linear, either in the realizable case or in the non-realizable one.
Thus, the set $C$ of predictors may be so large that no predictor can have a sublinear regret. We still would like to have as small regret as possible 
with respect to this set. Since the set $C$ is larger, the realizable case becomes more interesting. Can the smallest regret still be achieved with 
a Bayesian predictor? 
% 
%  \bibliographystyle{plain}
%  \bibliography{../my}

\begin{thebibliography}{1}

\bibitem{Cesa:06}
N.~Cesa-Bianchi and G.~Lugosi.
\newblock {\em Prediction, Learning, and Games}.
%\newblock Cambridge University Press, 2006.
\newblock CUP, 2006.
\bibitem{Diaconis:86}
P.~Diaconis and D.~Freedman.
\newblock On the consistency of {B}ayes estimates.
\newblock {\em Annals of Statistics}, 14(1):1--26, 1986.

\bibitem{Grunwald:07}
P.~Gr{\"u}nwald.
\newblock {\em The Minimum Description Length Principle}.
\newblock MIT Press, 2007.

\bibitem{Krichevsky:93}
R.~Krichevsky.
\newblock {\em Universal Compression and Retrival}.
\newblock Kluwer Academic Pub., 1993.

\bibitem{BRyabko:88}
B.~Ryabko.
\newblock Prediction of random sequences and universal coding.
\newblock {\em Problems of Information Transmission}, 24:87--96, 1988.

\bibitem{Ryabko:10pq3+}
D.~Ryabko.
\newblock On finding predictors for arbitrary families of processes.
\newblock {\em Journal of Machine Learning Research}, 11:581--602, 2010.

\bibitem{Ryabko:11pq4+}
D.~Ryabko.
\newblock On the relation between realizable and non-realizable cases of the
  sequence prediction problem.
%\newblock {\em Journal of Machine Learning Research}, 12:2161--2180, 2011.
\newblock {\em J. Machine Learning Research}, 12:2161--2180, 2011.
\bibitem{Solomonoff:78}
R.~J. Solomonoff.
\newblock Complexity-based induction systems: comparisons and convergence
  theorems.
\newblock {\em IEEE Trans. Information Theory}, IT-24:422--432, 1978.

\bibitem{Zvonkin:70}
A.~K. Zvonkin and L.~A. Levin.
\newblock The complexity of finite objects and the development of the concepts
  of information and randomness by means of the theory of algorithms.
\newblock {\em Russian Mathematical Surveys}, 25(6):83--124, 1970.

\end{thebibliography}
\subsection*{Acknowledgements}
{\small The research presented in this paper was supported by CPER Nord-Pas de Calais/ FEDER DATA Advanced
data science and technologies 2015-2020, by French Ministry
of Higher Education and Research, Nord-Pas-de-Calais Regional
Council.}

\end{document}